\documentclass[letterpaper, 10 pt, journal, twoside]{ieeetran}

\usepackage[noadjust]{cite}
\usepackage{amsmath,amssymb,amsfonts}
\usepackage{algorithmic}
\usepackage{subfigure}
\usepackage{graphicx}
\usepackage{textcomp}
\usepackage{xcolor}
\usepackage{soul}
\usepackage{graphics} 
\usepackage{epsfig} 
\usepackage{mathptmx} 
\usepackage{times} 
\usepackage{amsmath} 
\usepackage{amssymb}  
\usepackage{multirow}
\usepackage{graphicx, epstopdf}
\usepackage{xspace,epsfig,url}
\usepackage{psfrag}
\usepackage{verbatim}
\usepackage[all]{xy}
\usepackage{color}
\usepackage{comment}
\usepackage{cases}
\usepackage{hyperref}

\title{\LARGE \bf Perceived Intensities of Normal and Shear Skin Stimuli\\ using a Wearable Haptic Bracelet}
\author{Mine Sarac$^{1,4}$, Tae Myung Huh$^{2}$, Hojung Choi$^{1}$,\\ Mark R. Cutkosky$^{1}$, Massimiliano Di Luca$^{3}$, and Allison M. Okamura$^{1}$
\thanks{Manuscript received: September, 9, 2021; Revised: November, 24, 2021; Accepted: December, 19, 2021.}
\thanks{This paper was recommended for publication by Editor Jee-Hwan Ryu upon evaluation of the Associate Editor and Reviewers' comments. This work was supported in part by National Science Foundation grant 1830163, Combat Capabilities Development Command-Soldier Center (CCDC-SC) grant W81XWH-20-C-0008, and Facebook Reality Labs.}
\thanks{$^{1}$Stanford University, USA
        {\tt\footnotesize \{msarac, hjchoi92, cutkosky, aokamura\}@stanford.edu}}%
\thanks{$^{2}$University of California at Berkeley, USA, was at Stanford University, USA
        {\tt\footnotesize thuh@berkeley.edu}}%
\thanks{$^{3}$University of Birmingham, UK, was at Facebook Reality Labs, USA
        {\tt\footnotesize m.diluca@bham.ac.uk}}%
\thanks{$^{4}$Kadir Has University, Turkey
        {\tt\footnotesize mine.sarac@khas.edu.tr}}%
\thanks{Digital Object Identifier (DOI): see top of this page.}
}

\begin{document}

\maketitle

\begin{abstract}
Our aim is to provide effective interaction with virtual objects, despite the lack of co-location of virtual and real-world contacts, while taking advantage of relatively large skin area and ease of mounting on the forearm. We performed two human participant studies to determine the effects of haptic feedback in the normal and shear directions during virtual manipulation using haptic devices worn near the wrist. In the first study, participants performed significantly better while discriminating stiffness values of virtual objects when the feedback consisted of normal displacements compared to shear displacements. Participants also commented that they could detect normal cues much easier than shear, which motivated us to perform a second study to find the point of subjective equality (PSE) between normal and shear stimuli. Our results show that shear stimuli require a larger actuator displacement but less force than normal stimuli to achieve perceptual equality for our haptic bracelets. We found that normal and shear stimuli cannot be equalized through skin displacement nor the interaction forces across all users. Rather, a calibration method is needed to find the point of equality for each user where normal and shear stimuli create the same intensity on the user's skin. 
\end{abstract}

\section{Introduction}

\begin{figure}[t!]
  \centering
  \resizebox{3.4in}{!}{\includegraphics{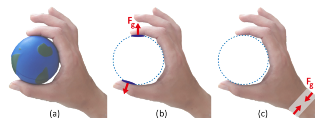}}
  \caption{Grasping tasks with different types of haptic feedback: (a) In the real world, the fingers directly contact the object. (b) In a virtual environment with fingertip haptic devices, grasp forces ($F_g$) are displayed on the fingertips. (c) In a virtual environment with a wearable haptic device, grasp forces are displayed on the forearm near the wrist.}
  \label{fig:proposed}
  \vspace*{-.5\baselineskip}
\end{figure}

In the real world, mechanical properties of objects, such as mass, stiffness, and temperature, can be perceived via touch (Fig.~\ref{fig:proposed}(a)). Haptic devices aim to recreate the same feeling for virtual interactions. Many multi-degree-of-freedom fingertip devices have been developed to render the interaction forces during active exploration/manipulation tasks in a virtual environment, as shown in Fig.~\ref{fig:proposed}(b)~\cite{Suchoski2018, Leonardis2017}. The combination of the shear and normal skin stimulation provided by these devices and the high density of mechanoreceptors in the fingerpad result in strong performance and perceived realism of manipulation tasks in virtual environments~\cite{Schorr2017}.

However, fingertip devices must be miniaturized to reduce encumbrance. Such a requirement complicates the design and increases the cost of actuators, which must have relatively large output force, small size and light weight. Furthermore, there are certain applications where fingertips should be left free to interact with physical objects, as during augmented reality. In these situations, users cannot wear fingertip devices which would interfere with these physical objects.

Here, we relocate the haptic stimulation from the fingertip to the arm. In doing so, the mechanical properties of manipulated virtual objects are rendered on the arm (Fig.~\ref{fig:proposed}(c)). In this context, haptic feedback is computed through fingertip contact and material properties of objects is rendered on the user's arm. We propose that haptic stimulation at or near the wrist that represents the real properties of an object, even if not rendering them perfectly, might be sufficient to create interpretable or ``believable" interactions. Such relocation could qualitatively add to (rather than detract from) the user experience without increasing cognitive or attentional load. 

Werarable bracelets and arm bands have been used to emulate the sensation of human touch in social interactions~\cite{Baumann2010, Wang2012, Culbertson2018}, map haptic cues to navigational directions~\cite{Nakamura2014, Chinello2018, Caswell2012} or communication~\cite{Dunkelberger2018, Song2015, Zheng2012}, render interaction forces during teleoperation tasks~\cite{Meli2018} or prosthesis control~\cite{Casini2015, Fripp2018, Tejeiro2012}, and improve the learning process for trainees in robotic surgical systems~\cite{Brown2017}. Tasbi~\cite{Pezent2019}, Bellowband~\cite{Young2019}, and HapWRAP~\cite{AghareseICRA2018} squeeze the wrist area in a distributed manner using various actuation methods.

Although wrist-worn devices have been shown to improve user performance during virtual manipulation tasks, there remain questions regarding the efficacy of different degrees of freedom of haptic feedback, namely in the normal (squeeze) and shear directions. (In this paper, we refer to devices mounted at or near the wrist as wrist-worn.) Moriyama et al.~\cite{Moriyama2018} developed a five-bar linkage mechanism with two degrees of freedom to present haptic feedback to the forearm during virtual interactions. They asked users to evaluate different feedback directions based on the ``strangeness'' feeling they create on users. Their results show that normal feedback feel less strange than shear when applied on the wrist. Despite their inspirational ideas regarding wearable wrist devices and perception, we found it difficult to map the strangeness metric toward practical design guidelines.

The effects of force direction on users' perception, task performance, and learning curve of virtual manipulation tasks for wrist-worn devices are still unknown. Previously, we developed haptic sketches by simulating interaction forces on user's arm manually~\cite{Sarac2019}. We applied normal forces and shear forces to user's wrist at the dorsal side as users interact with the virtual environment. Users reported that both normal and shear forces felt natural, intuitive, and interactive. 



In this paper, we first present the results of a study that compares the user performance and perception while discriminating virtual stiffness values of objects and receiving haptic feedback in the normal and shear directions acting on the wrist. The stiffness discrimination study was previously designed and presented as an extended abstract~\cite{Sarac2020}; here we fully describe the study and results. Then we investigate the difference between the perceived intensity of normal and shear stimuli in terms of actuator displacements and applied forces. Finally, we propose a perceptual model of shear displacement intensity with respect to normal.

\begin{figure}[b!]
  \begin{center}
  \vspace*{-.5\baselineskip}
\includegraphics[width=\columnwidth]{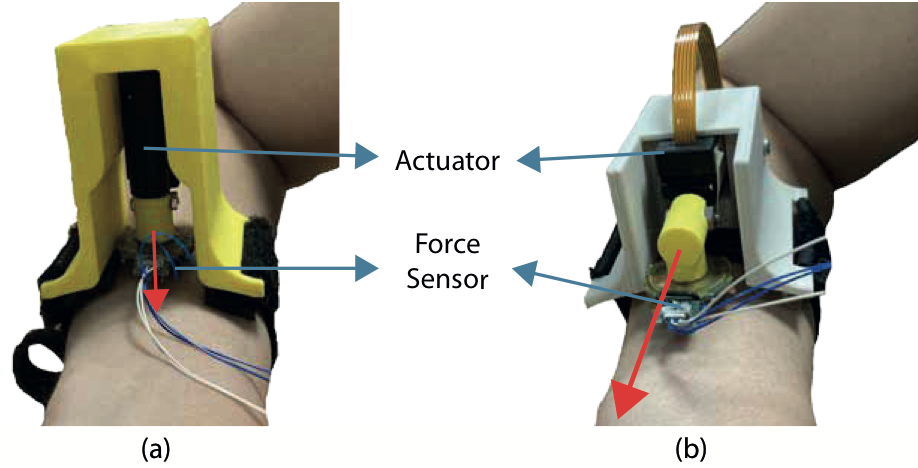}
  \end{center}
  \caption{Haptic bracelets worn by a user on the wrist provide skin deformation as (a) normal stimulus and (b) shear stimulus. Double-sided tape between the end-effector and the skin prevents slip during shear stimulus. The force sensor is used only for Experiment 2.}
    \label{fig:direction}
\end{figure}

\section{Haptic Bracelets}

To study the effect of force direction (normal versus shear) on perception during virtual interaction, we used Actuonix PQ12-P linear actuators due to their weight (15~g), maximum stroke (20~mm), high output force (18~N), and easy controllability via an integrated position sensor. The haptic bracelet is composed of two actuator sets on the dorsal and ventral sides of the forearm, and weighs less than 40 g.


We selected grounding/orienting of linear actuators to enable investigation of both direction and location of forces acting around the wrist. The direction of forces is adjusted by designing different grounding parts as shown in Fig.~\ref{fig:direction}. Grounding the linear actuator vertically on the wrist applies normal forces as the displacement is controlled (Fig.~\ref{fig:direction}(a)). Grounding the actuator horizontally creates shear forces with double-sided tape used to prevent the end-effector from slipping on the skin (Fig.~\ref{fig:direction}(b)). The bracelets are designed to be worn on the user's forearm near the wrist to minimize the impact of wrist movements and facilitate consistent physical contact. The grounding is designed with a curvature to fit the forearm, a silicone pad between the plastic and the skin, and wide Velcro straps to keep the grounding stable.


\section{Experiment 1: \\ Stiffness Discrimination} \label{sec:exp1}

\begin{figure}[t!]
  \centering
  \vspace*{.5\baselineskip}
  \resizebox{3in}{!}{\includegraphics{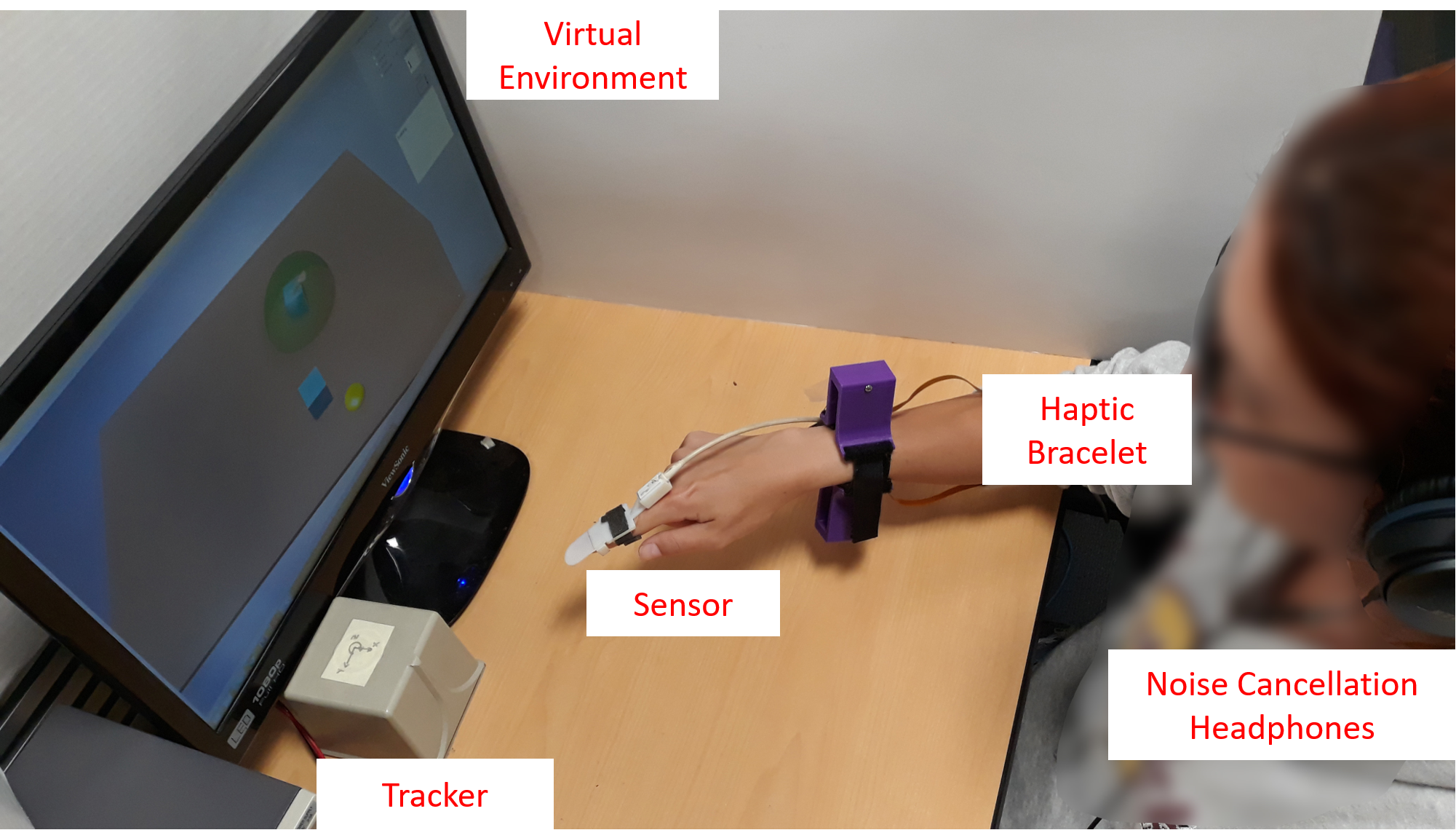}}
  \caption{Set up for Experiment 1: A user sits in front of a monitor and wears a haptic bracelet, a fingertip tracking sensor, and noise cancellation headphones. Users are asked to interact with objects in the virtual environment while the haptic device renders the interaction forces at the wrist.}
  \label{fig:setup1}
  \vspace*{-.5\baselineskip}
\end{figure}

To identify the direction in which interaction forces should be applied to the arm to improve the perception and performance of virtual tasks, we performed a stiffness discrimination  experiment using the haptic bracelets, a virtual environment, and tracking system as shown in Fig.~\ref{fig:setup1}.

12 participants (age 24-31, 6 females and 6 males) joined the study. All participants were right-handed and had previous experience with haptic interfaces. The Stanford University Institutional Review Board approved the experimental protocol and all participants gave informed consent. 

\subsection{Experiment Procedures}

\begin{figure*}[]
\centering
{\includegraphics[width=0.87\textwidth]{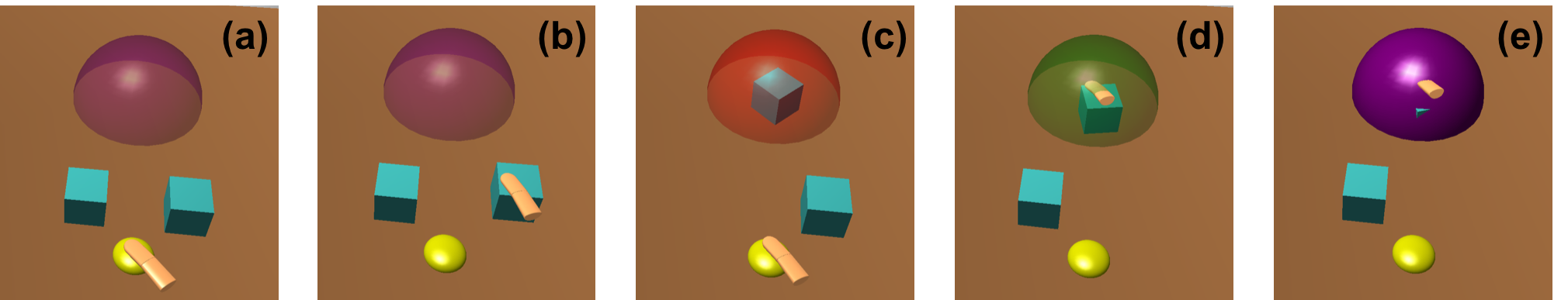}} 
\caption{Experiment 1 task: (a) Each trial has a starting zone, target zone, and two visually identical objects with different stiffness values. (b) The participant presses on each object, chooses the stiffer object based on the haptic feedback, and drags it to the purple zone. During training, the zone turns (c) red if the answer is wrong, and (d) green If the answer is correct. (e) During testing, the zone turns purple and opaque regardless the answer.}
	\label{fig:TrainingScheme}
\vspace*{-.5\baselineskip}
\end{figure*}

We created a virtual environment using the CHAI3D framework~\cite{Conti2005} as shown in Fig.~\ref{fig:TrainingScheme}. During the experiments, the virtual environment is displayed on a regular monitor and updated at 144~Hz. User's finger movements are tracked at approximately 200~Hz using  a trakSTAR tracking system and an Ascension Model 800 sensor attached on user's finger through 3D printed grounding.

The overall experiment is composed of two parts, one for each direction of force (normal vs. shear). For each part, there is a training block with 24 trials and 3 testing blocks with 16 trials each. Once the first part is completed, the participant wears the bracelet with the other force direction and repeats the entire procedure. Between each block and part, the participant was given a break time to rest as needed. Each participant performed the task in a different order.

There are two objects in the virtual environment. The stiffness value of one object is kept at 300~N/m. and the other object is pseudo-randomized among 100, 200, 400, and 500~N/m. Their initial locations are randomized.


During the experiment, the participant sees two identical boxes which have different simulated stiffness values. Participants move their index finger so that their avatar in the virtual environment interacts with these virtual objects. Specifically, they press on each object to evaluate its stiffness and drag the stiffer object to the target zone. While interacting with the virtual objects, the participants were not constrained in terms of their interaction strategies or finger movements. The experiment has two modes: training and testing. The training mode aims to guide the user towards interpreting the haptic cues correctly while assessing the stiffness values of the virtual boxes. Therefore, the target zone changes color based on the participant's answer. If the answer is correct, the zone turns green, and if the answer is wrong, the zone turns red (Figure \ref{fig:TrainingScheme} (c) and (d)). During the testing mode, the target zone becomes opaque to indicate task completion. 

As the participant interacts with virtual boxes, rendered forces are computed based on the stiffness values. The linear actuators are position controlled, so desired forces are expressed in the form of desired displacements using a fixed force-to-displacement ratio 0.03~N/mm~\cite{Diller2001}. Even though hairy skin was previously reported with stiffness of 0.03~N/mm for normal and 0.04~N/mm for shear directions, whether skin stiffness is different for the dorsal and ventral sides of the wrist is unknown. Thus, in order to provide consistent stimuli, we kept the level of displacements (as opposed to force) the same for the different conditions. Biggs et al. showed that hairy skin is three times more sensitive to shear displacements than normal~\cite{Biggs2002}, so we foresee that stiffness discrimination would be better with shear feedback.

\subsection{Results and Discussion}

\begin{figure}[]
\centering
{\includegraphics[width=0.48\textwidth]{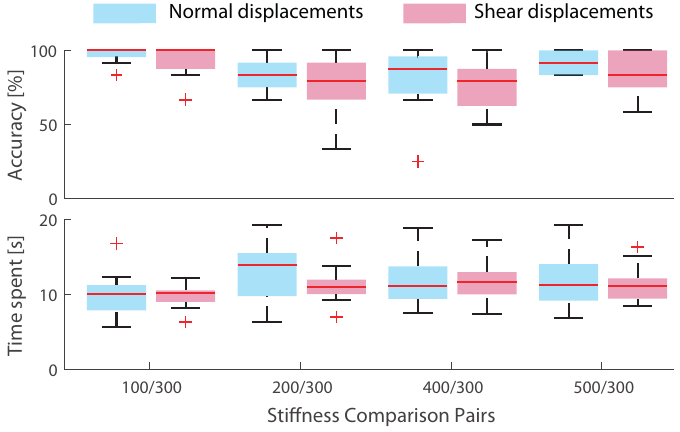}} 
\vspace*{-.5\baselineskip}
\caption{Task accuracy and the time spent on trials for Experiment 1. The main effect of the feedback direction was significant in terms of task accuracy, but not time spent.}
	\label{fig:DirectionComparisonOverStiffness}
	\vspace*{-.5\baselineskip}
\end{figure}

We investigated the task accuracy and time spent across all participant as shown in Fig.~\ref{fig:DirectionComparisonOverStiffness}. We compared participants' accuracy using a two-way repeated-measure  analysis  of  variance (ANOVA) with factors of direction of force feedback (normal or shear) and  stiffness value. Mauchly's test ($\chi^2(5) = 5.75, p = 0.33$) did not indicate a violation of sphericity. The main effect of feedback direction was significant ($F(1,88)=5.831, p=0.018$, $\eta^2_{partial}= 0.062$), indicating that participants performed better using normal feedback than shear. In addition, there was an influence of compared stiffness pairs ($F(3,88) = 4.596, p=0.005$, $\eta^2_{partial}= 0.135$). The interaction effect was not significant ($F(3,88) = 0.733, p = 0.537$, $\eta^2_{partial}= 0.009$). We also compared the average time spent to complete the trials using a two-way repeated-measure ANOVA with the same factors. The main effect of the stiffness pair was significant ($F(3,88) =  2.717, p=0.049$, $\eta^2_{partial}= 0.085$). However, we found no significance for the main effect of feedback direction ($F(1,88) =  0.721, p=0.398$, $\eta^2_{partial}= 0.008$) or for interaction $F(3,88) =  0.883, p=0.219$, $\eta^2_{partial}= 0.007$).

In addition, we collected subjective comments from participants. When asked the direction of displacement they liked the most, four users chose shear and eight chose normal. However, when asked which feedback direction was easiest to notice, all participants answered normal. These subjective comments are coherent with the analyses performed above, so we conclude that normal forces are more effective than shear for stiffness recognition in this setup. 

Our results contradicted those of Biggs et al.~\cite{Biggs2002} regarding sensitivity of normal versus shear force on the skin. This contradiction and the verbal comments from the participants led us to believe that equalizing displacements using wrist-worn devices created a bias in perception, especially while comparing normal and shear feedback directions. One possible solution is to use force measurements to control the interaction forces rather than the displacements. However, when we measured the interaction forces while actuators provided controlled displacement levels, we realized that the wrist movements and muscle activity affected the force measurements in multiple degrees of freedom. Since the actuators have only one degree of freedom, these interaction forces cannot be fully controlled. Therefore, we performed a second experiment to investigate strategies for equalization.


\newpage
\section{Experiment 2:\\ Point of Subjective Equality}

\begin{figure} [b!]
     \centering
     \vspace*{1\baselineskip}
     \resizebox{3.2in}{!}{\includegraphics{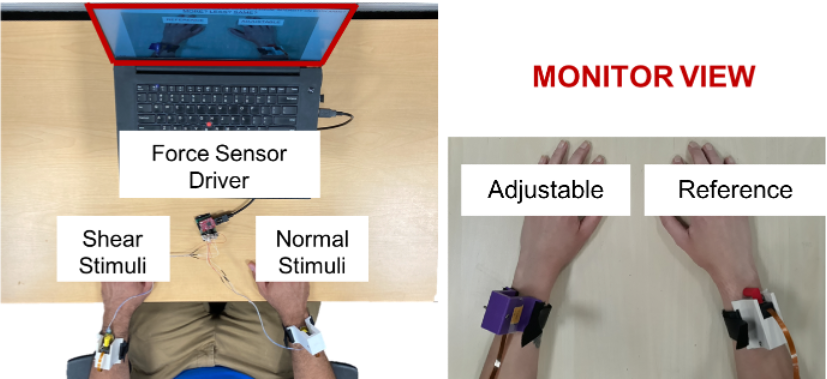}} 
     \caption{Setup for Experiment 2: A participant wears two bracelets with normal and shear stimuli. A monitor gives a visual instruction of which arm receives the reference and adjusted stimuli.}     
     \label{fig:setup}
\end{figure}

In this experiment, we investigate and quantify differences between the perceived intensity of normal and shear stimuli on the wrist applied by a wearable haptic bracelet. Participants perceive normal and shear stimuli on separate wrists and tune the intensities until both feel the same, as shown in Figure \ref{fig:setup}. We determined the relative intensity of normal and shear stimuli based on actuator displacements, how the perceived intensity of normal and shear stimuli are affected by applied force, whether the point of subjective equality varies across people, and how perception of shear displacement can be modeled with respect to normal displacements.

For this experiment, the haptic bracelets are equipped with 6-degree-of-freedom (6-DoF) force sensors to measure interaction forces and equalize the tightness of the Velcro strap on each bracelet. The custom capacitive force sensors are similar to \cite{Wu2015}, have a compact form (2 cm in diameter, 3.2 mm thickness), and are low-cost (\textless\$10) and high performance compared to commercial sensors. The RMS errors between the data collected from the custom force sensor and a commercial ATI Gamma force sensor were 0.0752~N and 0.0617~N when calibrated in the range of $<5~N$ and $<6~N$ in the shear and normal directions, respectively. Our measurements confirmed that the two bracelets generated the same amount of force in their respective directions of actuation, and the reaction force on the other side of the wrist was undetectable via the force sensor due to the distribution of the reaction force over the large contact area of the belt.


In addition to the experiment described in Section~\ref{sec:exp1}, two prior works provided motivation for the design of this study. 
Diller et al. \cite{Diller2001} performed a study using a grounded, flat-ended probe on the skin. 
Their results showed that shear forces were less than normal forces for the same actuator displacements, but how participants perceived these stimuli is not reported. As described earlier, Biggs et al.~\cite{Biggs2002} compared normal and shear stimuli at the forearm in terms of perceived intensity, and found that shear displacement should be 3 times less than normal displacement in order to create a similar intensity. 
We hypothesize that differences in experimental setup (wearability, contact area, reference stimulus value) might change the relative actuator displacements for normal and shear stimuli, but not the interaction forces.


Four participants (age 25-32, 1 female and 3 males) and eight participants (age 25-32, 3 females and 5 males) joined the preliminary and main studies, respectively. Participants of the preliminary study also participated to the main study. All participants were right-handed and had previous experience with haptic interfaces. The Stanford University Institutional Review Board approved the experimental protocol and all participants gave informed consent. 

\subsection{Experiment Procedures}

Upon the participant's arrival, the experimenter spent 5 minutes adjusting the setup and explaining the experiment procedures. Participants were seated on a chair with their elbows supported by the chair and their hands by a desk, so that their wrists were suspended and contacted only the haptic bracelets. 
Participants were asked to maintain this posture throughout the experiment to avoid muscular activity on the forearm; when the experimenter visually observed a change, the participants were asked to re-adjust their posture. 

The experimenter secured the bracelets to the forearm using two Velcro straps, following a procedure to match the tightness on the two arms. The experimenter first fixed the Velcro of one bracelet at a tightness sufficient to secure the bracelet, but not so tight as to cause discomfort. After the first bracelet was secured, the experimenter checked the marks on the Velcro straps (at 0.5~mm increments), and initially fixed the second bracelet at the same marks. After both Velcro straps were fixed on the second bracelet, the force measurements were checked and the Velcro of the second bracelet was adjusted if they differed of more than 0.1~N. 


The participants wore headphones with white noise to reduce the environmental noise and experimenter's instructions. This experiment was conducted during the COVID-19 pandemic with safety precautions such as social distancing, disinfecting the setup between each uses, and face masks. 

\subsubsection{Task}

\begin{figure}[]
     \centering
     \vspace*{1\baselineskip}
     \resizebox{3.1in}{!}{\includegraphics{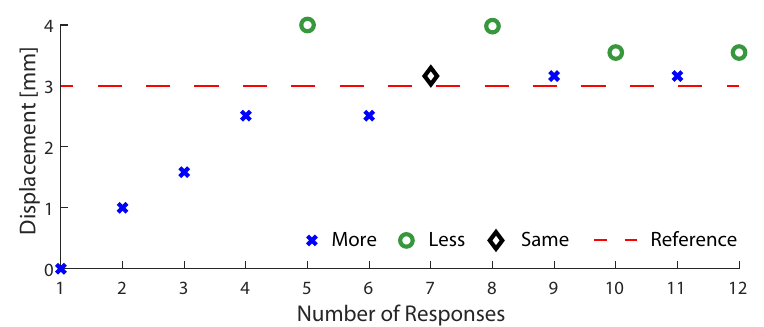}}
     \vspace*{-.5\baselineskip}
     \caption{Example data from one participant, demonstrating convergence via the method of adjustments. The staircase algorithm calculates actuator displacements based on the participant's responses. The data was collected during the main study of Experiment 2 with 3 mm shear reference.}
     \label{fig:staircase}
     \vspace*{-1\baselineskip}
\end{figure}



The experiment consisted of a number of trials where the participant verbally reported whether one of the stimuli felt more, less, or equally intense to the reference despite the difference in the actuation direction. Participants were indicated on a monitor which wrist received the fixed reference stimulus and which stimulus would be adjusted according to their response. After experiencing each stimulus pair, participants reported whether the adjusted stimulus should be increased  (response `more'), if it should be decreased  (response `less') or 
if it felt the same as the reference  (response `same'). 
Based on the participant's response, the actuator displacement for the adjusted stimulus was computed for the next trial using the staircase method and ``two-up one-down'' paradigm~\cite{Levitt}. Fig.~\ref{fig:staircase} shows an example trial with participant responses and actuator displacement of the adjusted stimulus. For iteration $n$, if the participant's response was `more', the actuator displacement was computed as: 

\vspace*{-0.5\baselineskip}
\begin{equation}
    X(n) = X(n-1)*(10)^{(Ldb/20)}
\end{equation}
\vspace*{-1.0\baselineskip}

\noindent where $X(n)$ and $X(n-1)$ are the current and previous displacements and $Lbd=4/(2^i)$ is the stimulus intensity. $i$ is how many times the participant gave the `less' response, such that with each `less' response, the actuator displacement returned to the previous value ($X(n)=X(n-2)$) and $i$ was increased by 1. When participants gave the response `same', the algorithm computed the next iteration using the previous response in the sequence. Thus,  participants can express to have perceived equal stimuli, while allowing the staircase algorithm to overshoot the levels where both stimuli feel equally intense. The staircase was terminated if for the last 10 trials the average displacement change was less than 0.5~mm.


After each response, both the reference and the adjusted stimuli return to their neutral states (0~mm actuator stroke). To minimize the impact of actuation velocity or duration differences in duration, reference and adjusted stimuli were applied in sequence, as first the reference and the adjusted stimuli with 2-second pause in between. The experimenter verbally announced the reference and adjusted stimuli before actuation. The shear displacement reference less than 3 mm was not used in the experiments because volunteers for the pilot testing reported that they did not feel confident to compare the intensities between normal and shear stimuli even though they perceived the signals rendered on their skin. On the other hand, normal stimuli of more than 3 mm caused discomfort. Therefore, 3 mm was found to be an effective reference value for both stimuli.

\subsubsection{Preliminary Study: Comparing the Same Stimuli}

Before conducting the main study to compare normal and shear stimuli, the proposed experiment and the method of adjustments were validated with a preliminary experiment, where participants wore haptic bracelets rendering the same type of stimuli on both wrists (normal-normal or shear-shear) and compared the intensity of the same stimuli. Half of the participants compared the normal stimuli first and the shear stimuli second, while the other half compared the shear stimuli first and the normal stimuli second.

For each participant, 1 training trial and 4 testing trials were performed with the first set of bracelets: the reference was given on the right arm during 2 testing trials and on the left arm during 2 testing trials. The order of the trials with reference stimulus on the right or left arm were changed for each user as well. Once the participant completed 4 testing trials, the same procedure was repeated with the second set of bracelets: normal bracelets were replaced with shear, and shear bracelets were replaced with normal. During all trials, the reference actuator displacement was set as 3 mm.  

\subsubsection{Main Study: Comparing Different Stimuli}

\begin{table}[t!]
\centering
\vspace*{1\baselineskip}
\resizebox{3.3in}{!}{\includegraphics{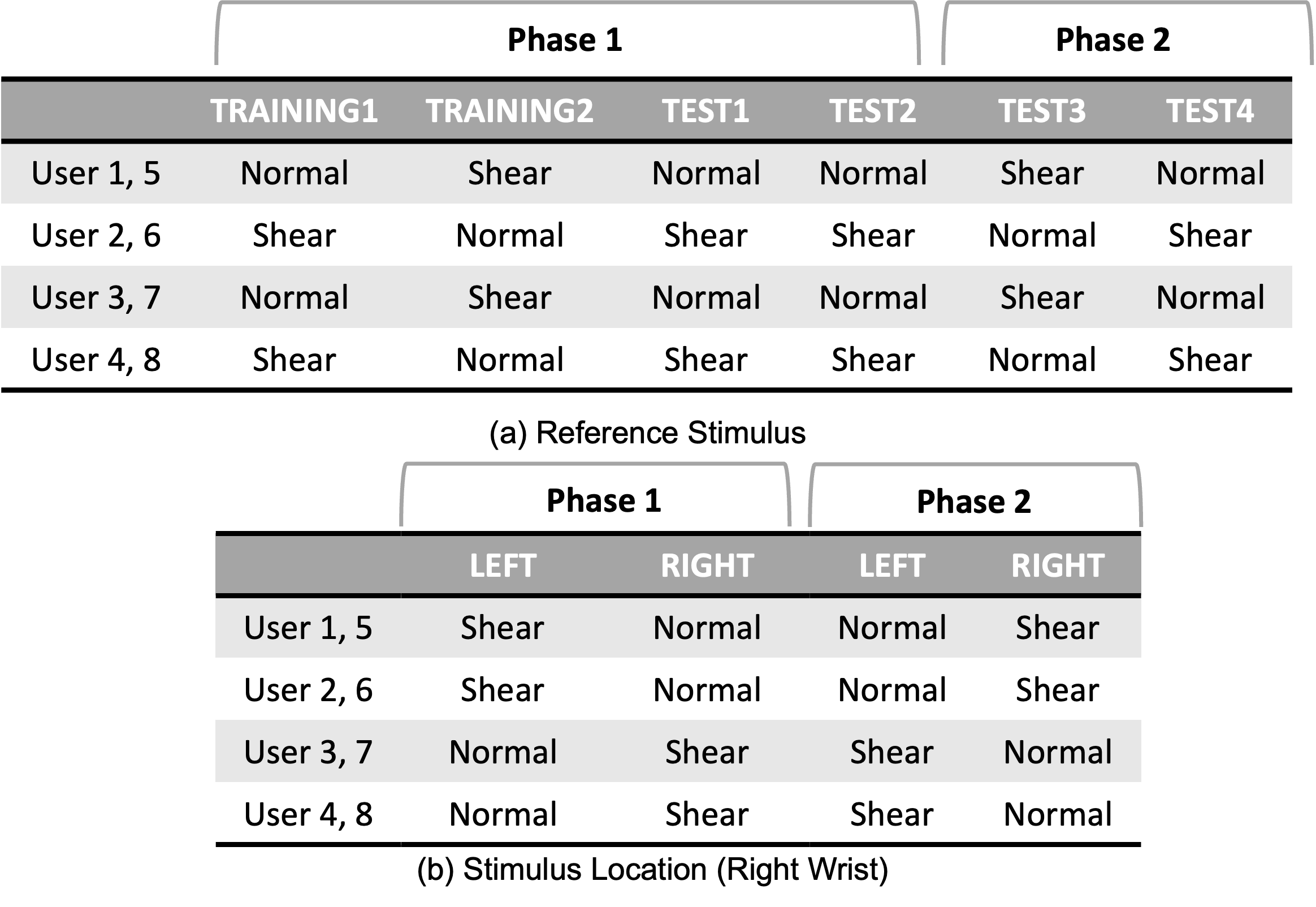}} 
\caption{Protocol and flow of Experiment 2 for each user}
\label{tab:protocol}
\vspace*{-2\baselineskip}
\end{table}

\begin{figure*}[t!]
     \centering
\vspace*{1\baselineskip}
         \resizebox{7in}{!}{\includegraphics{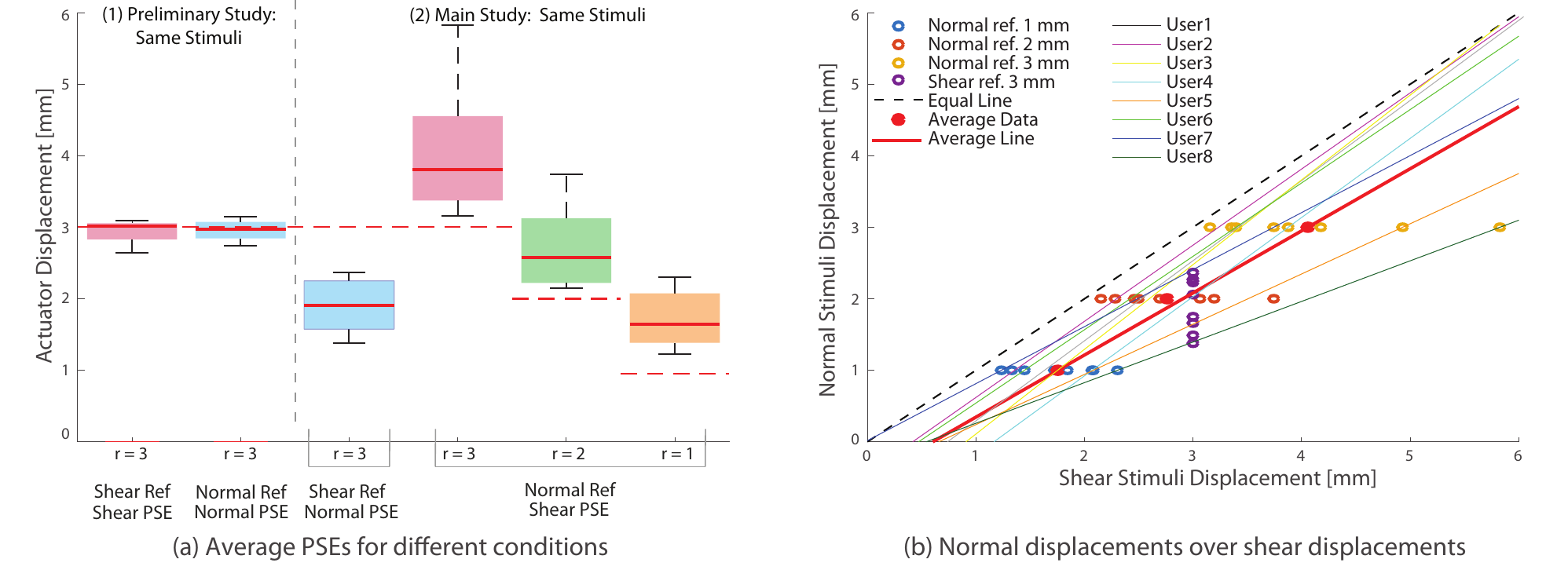}} 
         \vspace*{-.5\baselineskip}
     \caption{Results of Experiment 2, for point of subjective equality: (a) PSE of actuator displacements with (1) preliminary and (2) main studies. Red dashed lines show the reference values, blue boxes indicate normal PSEs, red boxes indicate shear PSEs with 3~mm reference, green box indicates shear PSEs with 2~mm reference, and orange box indicates shear PSEs with 1~mm reference. ‘r’ represents the actuator displacement for the reference stimulus. (b) A fit line model between actuator displacements of shear PSE and normal reference for each user. Labels are ordered based on the line slope. The results indicate that a larger displacement in the shear direction is needed by all participants to equalize its intensity with respect to the normal direction.}
              \label{fig:AvgPSE}
          \vspace*{-1\baselineskip}
\end{figure*}

Table~\ref{tab:protocol} shows the experiment flow with two main phases: in Phase~1, the participant wore the bracelets as designated and performed two training and two testing blocks. In Phase~2, the location of the bracelets were switched and the participant performed two more testing blocks. Participants familiarized themselves with the task during two training blocks. Each training block included one trial with a 3 mm normal reference and one trial with a 3 mm shear reference. During the training blocks, the participants were encouraged to ask questions about the task. After the training blocks were performed, the experimenter made sure that all participants were confident with the task. Then, the testing blocks started. 

Pilot tests revealed that the shear reference cannot be less than 3 mm to be noticed easily. Consequently, the participants compared normal and shear stimuli with (i) 1, 2, and 3 mm normal reference and (ii) 3 mm shear reference. The participants repeated each reference value 3 times. The order of the reference values were randomized for each testing block with normal reference. Participants received a 1-minute break after each set of 6 testing trials. Between Phase~1 and Phase~2, participants swapped the locations of bracelets (exchanging which arm received normal and shear stimuli), thus they received a longer break time of approximately 5 minutes including the preparation time.

\subsection{Results} \label{sec:results}

\subsubsection{Preliminary Study: Comparing the Same Stimuli}

Fig.~\ref{fig:AvgPSE}(a,1) shows the average PSEs in terms of actuator displacements when the participants compared the same stimuli (normal-normal and shear-shear). One-tailed t-tests indicate that the mean PSEs are not statistically different for the  normal stimulus adjusted to match the normal reference ($t(15)=-0.6884, d = -1.7216, p = 0.517$) and for the  shear stimulus adjusted to match the shear reference ($t(15)=-1.0121, d = -0.2531, p = 0.327$). These results indicate the validity of the experiment and the ability of participants to compare stimuli intensities acting on the wrists.

\subsubsection{Main Study: Comparing Different Stimuli}    

\textbf{Actuator Displacements:} PSEs were analyzed in terms of actuator displacements (mm), as shown in Fig.~\ref{fig:AvgPSE}(a,2). Each bar represents the average PSE, and the horizontal dashed lines show the reference value corresponding to each bar. The first (blue) bar shows the average normal PSE (1.9 mm) for 3 mm shear reference. The second (red) bar shows the average shear PSE (3.7 mm) for 3 mm normal reference. The third (green) bar shows the average shear PSE (2.7 mm) for 2 mm normal reference. Finally, the forth (orange) bar shows the average shear PSE (1.7 mm) for 1 mm normal reference.  

Each participant performed 3 repetitions for each condition, and the PSEs were averaged and submitted to a two-way repeated-measure ANOVA with factors reference stimulus (1, 2, and 3 mm normal and 3 mm shear) and which arm was the reference applied to (dominant or non-dominant side). Mauchly’s test ($\chi^2(5) = 1.991, p = 0.852$) did not indicate a violation of sphericity. There was an influence of the reference stimulus and type of reference on PSE ($F(3,56) = 42.994, p<0.001$, $\eta^2_{partial}= 0.697$). Most notably, single-sample t-tests indicate a difference between PSE and the reference displacement for all four conditions ($p < 0.01$). A post-hoc Tukey test showed that there was no difference in PSE between 1 mm normal and 3 mm shear references stimuli, while all the other paired comparisons were significant, see Fig.~\ref{fig:AvgPSE}(right). The main effect of the arm dominance was not significant ($F(1,56)=1.197, p=0.279$, $\eta^2_{partial}= 0.021$), indicating no consistent change in PSE whether the reference is applied to the dominant or non-dominant arm. The interaction effect was also not significant ($F(3,56) = 0.733, p = 0.537$, $\eta^2_{partial}= 0.038$), implying that the reference has the same effect on both arms.

Fig.~\ref{fig:AvgPSE}(b) shows the relationship between shear PSEs over normal reference. Data for each participant and on average were modeled with a linear fit. We calculated a Pearson coefficient to observe the correlation between the data collected for each reference stimulus. We have found no correlation between participants for all reference stimulus, implying that all participants perceived the stimuli differently.

\begin{figure*}
  \centering
  \resizebox{6.8in}{!}{\includegraphics{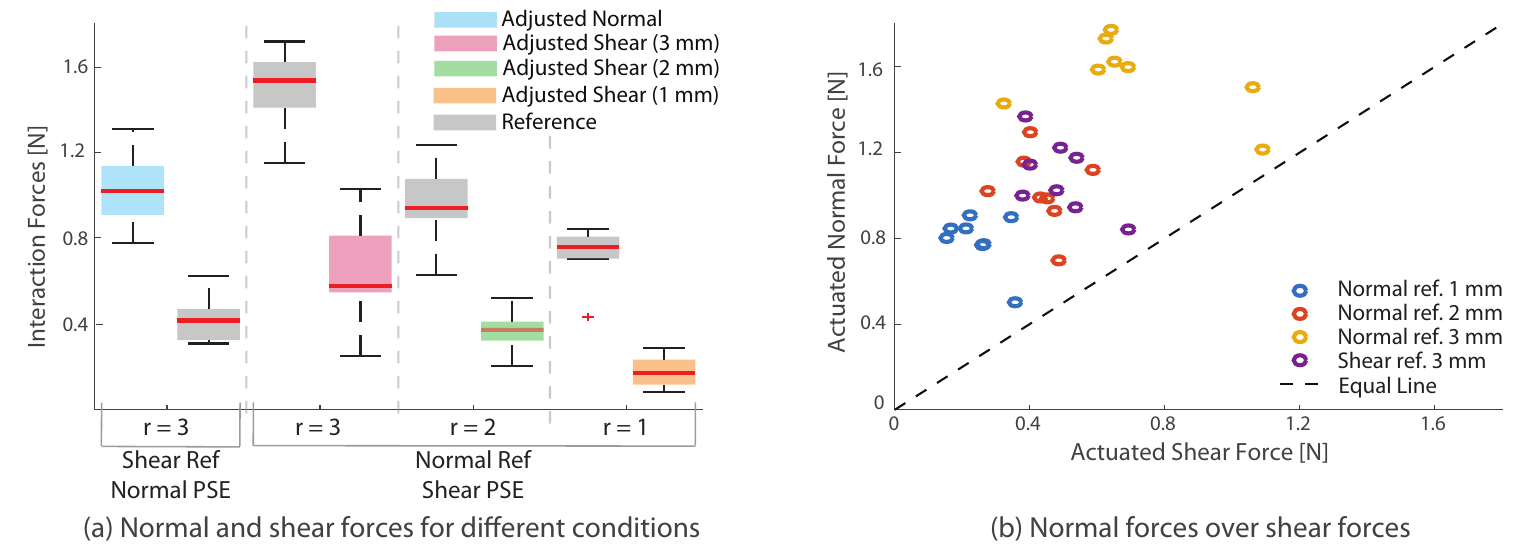}}
  \vspace*{-.500\baselineskip}
  \caption{Measured interaction forces for Experiment 2 when the normal reference is given as 1, 2, 3 mm and shear reference is given as 3 mm. ‘r’ represents the actuator displacement for the reference stimulus. The results indicate that normal stimuli create larger interaction forces than shear stimuli when their intensities are equalized.} 
  \vspace*{-1\baselineskip}
 \label{fig:AvgForces}
\end{figure*}

\textbf{Interaction Forces:} While participants were asked to perceptually match the sensation produced by the two types of displacement, it is possible to analyse the stimuli also in terms of the interaction \textit{forces}. The recordings indicate that the forces in the non-actuated directions do not vary across trials, so only the forces in the actuated direction are discussed here. Fig.~\ref{fig:AvgForces} shows the interaction forces recorded at the end of each trial during different references and averaged for all trials and participants. 


A two-way ANOVA on the influence of two independent variables (stimuli type and reference) was conducted on the measured interaction forces. All effects were statistically significant at the 0.05 significance level. The main effect for stimuli type (normal vs.\ shear) yielded an $F$ ratio of $F(1,56)=271.798, p<0.001, \eta^2_{partial} = 0.829$, indicating that when participants thought that normal and shear stimuli intensities were equal, the interaction forces were statistically significantly different from each other. Similarly, the main effect for reference type gave an $F$ ratio of $F(3,56)=42.389, p<0.001, \eta^2_{partial} = 0.694$. The interaction effect was found as $F(3,56)=3.033, p<0.037, \eta^2_{partial} = 0.140$. 
Interaction forces with normal and shear stimuli were statistically significantly different for all reference displacements. 

A post hoc Tukey test showed that the PSE of interaction forces collected with the 2 mm normal reference and 3 mm shear reference were not statistically significantly different from each other, while all the other comparisons were statistically significantly different. For 2 mm normal reference and 3 mm shear reference, the PSE of actuator displacements on average were calculated as 2.76 mm and 1.9 mm, respectively. While the ANOVA results indicate that the PSEs of actuator displacements are statistically significantly different from each other, their average is somewhat similar. It is possible that such difference is caused by the fact that the PSEs of actuator displacements vary across people. So, even though on average the participants reached the same interaction forces with similar actuator displacements, how they individually perceive the applied forces can vary. 

\subsection{Discussion}

\textbf{Relative intensity of normal and shear stimuli based on actuator displacements.} 
The PSEs showed that normal and shear stimuli can create the same intensity when shear stimuli had a larger displacement than normal (Fig.~\ref{fig:AvgPSE}(a,2)), regardless of participant, reference stimulus type, reference stimulus intensity, or which wrist received the reference. 

These results support the subjective comments received from the study in Section \ref{sec:exp1}, but contradict the findings previously presented by Biggs~et al.~\cite{Biggs2002}. Since the main purpose of this experiment has been to support the subjective comments received from our wearable haptic bracelet, the main reasons for this contradiction has not been investigated deeply. Nevertheless, we believe that these reasons can be the differences between the actuator base or the probe diameter.  


\textbf{How the perceived intensity of normal and shear stimuli are affected by applied force.} The force measurements (Fig.~\ref{fig:AvgForces}) showed that normal and shear stimuli can create similar intensities with different forces. For each participant, normal stimuli resulted in statistically significantly greater forces than shear stimulus needed to create a similar perceived intensity at the wrist. In other words, participants were more sensitive to shear forces compared to normal forces. Unlike the actuator displacements, interaction forces measured in this study show a similar trend to the findings of Biggs~\textit{et al.}~\cite{Biggs2002}. (They estimated forces using the actuator displacements they collected from the participants and skin stiffness measurements reported by Diller \textit{et al.}~\cite{Diller2001}.) This indicates that participants can perceive force stimuli similarly, regardless of how the haptic device is grounded.

\textbf{Does the point of subjective equality vary across people?}
ANOVA results with the main effect of participants indicate that the PSEs of actuator displacements are statistically significantly different for each participant. Even though some participants performed fairly similar to each other (see Fig.~\ref{fig:AvgPSE}(b)), claiming a constant ratio between normal and shear stimuli for all users would be erroneous. Instead, the relationship between normal and shear stimuli should be obtained for each user with a calibration process before using the wearable haptic devices to perform virtual or telemanipulated tasks where the relative perception of normal and shear stimuli are important.

\textbf{How perception of shear displacement intensity can be modeled with respect to normal displacements.} Our results showed that PSE of actuator displacements with normal stimuli is much lower than shear stimuli, while interaction forces with normal stimuli is much greater than shear stimuli. This means that equalizing normal and shear stimuli in terms of actuator displacements or interaction forces would create a perceptual bias if, for example, normal and shear stimuli were used as sensory substitution for the same stimulus. In this study, we implemented a linear model (Fig.~\ref{fig:AvgPSE}(b)) using three data points collected with different values of normal reference; more data points could be collected to obtain a more accurate relationship. We found that an offset displacement is needed for shear stimulus to create the same intensity when the normal stimulus is in the rest state (0 actuator displacement). This offset might be caused by how we secure the bracelets on the wrist. Both normal and shear bracelets are attached to the wrist through straps, which squeeze the wrist. Even though the normal stimuli applied on the wrist is perceived at a single point rather than the whole wrist, it is possible that the strap adds to the actuated normal stimulus given during the trials.

\section{Conclusions}

In this work, we analyzed the effects of haptic feedback directions rendered on user's forearm based on virtual interactions. Our results showed that normal displacements enabled participants to differentiate different stiffness levels significantly better than shear -- especially when the difference between virtual object stiffness values are large. Even though all participants reported that haptic cues are easier to notice with normal displacements, some reported that they enjoyed shear more because it was more ``subtle and calm'' compared to the normal forces. We hypothesize that participants favor shear displacements because shear skin stimulation is more closely related to pleasant stroking sensations \cite{Olausson2002}.
The second study measured point of subjective equality and investigated how to equalize normal and shear stimuli applied by a wearable haptic bracelet. 

Our findings show that normal and shear stimuli should be equalized with a calibration for each participant in terms of stimuli intensity. In future work, we will compare task performance of virtual manipulation while users perceive normal and/or shear haptic feedback. Such a comparison is possible only after we implement a calibration process based on the experimental procedure we presented in this work.

We will continue exploring the design and perception of wearable haptic devices. Beyond the wrist, mounting a haptic device at different body locations may have advantages, but presents challenges for secure attachment to the body and distribution of reaction forces \cite{CulbertsonAR2018}. In addition, it is important in the design process to understand how mechanoreceptors differ on different parts of the body \cite{ChenTOH2016} and how movement of the body (e.g., walking) can affect perception \cite{ShullTOH2019}.





\section*{ACKNOWLEDGMENT}
The authors thank collaborators at Facebook Reality Labs and Triton Systems, Inc. for their guidance on this work.


\bibliographystyle{IEEEtran}
\bibliography{HapticBracelet_ICRAwRAL}

\end{document}